\documentclass[runningheads]{llncs}
\usepackage[T1]{fontenc}
\usepackage{graphicx}
\usepackage{tabularx}
\usepackage{float}
\usepackage{booktabs}
\usepackage{amsmath}
\usepackage{svg}

\usepackage{tikz}
\usepackage{listings}
\usetikzlibrary{shapes.geometric, arrows.meta, positioning, fit}
\begin{document}
%
\title{Evaluating Structured Information Extraction with Open Models in a High Risk Public Sector Application}
\titlerunning{Evaluating Open Structured Information Extraction}
%
\author{Elias Schubert\inst{1} \and
Felix Bießmann\inst{1,2}\orcidID{0000-0002-3422-1026}}


%
\authorrunning{Schubert and Bießmann}
%
\institute{Berlin University of Applied Sciences, Luxemburger Str. 10, 13353 Berlin, Germany\\
\email{\{s92538,felix.biessmann\}@bht-berlin.de}\\
\url{https://www.bht-berlin.de}\and
Einstein Center Digital Future, Wilhelmstraße 67, 10117 Berlin, Germany\\
\url{https://www.digital-future.berlin}}

\maketitle     

\begin{abstract}

The extraction of structured information from unstructured documents represents a critical component of digital transformations in all sectors.  While proprietary solutions dominate commercial applications, a rapidly growing ecosystem of open-source Optical Character Recognition (OCR) engines, Large Language Models (LLMs), and Vision-Language Models (VLMs) offers accessible alternatives. However, systematic evaluations on realistic, multi-step extraction pipelines remain scarce. 
Responsible usage of such extraction tools require comprehensive evaluations on realistic tasks, especially as these solutions will be key components of applications in the public sector that the EU AI act categorizes as high risk. To address this gap we present a comprehensive benchmark assessing the end-to-end performance of open-source systems on a complex real-world document processing task classified as high risk: Student applications for an international study program. We conduct a comprehensive empirical evaluation with state-of-the-art OCR engines, LLMs and VLMs. Our results reveal that while VLMs generally outperform OCR+LLM pipelines, even state-of-the-art open-source models struggle to handle such tasks reliably in zero-shot settings. Only 4 of 35 configurations achieved F1 scores above 0.5, with the best OCR+LLM pipeline matching top VLM performance, though most OCR+LLM combinations performed substantially worse. Roughly 75\% of all configurations scored below 0.25. Model scale influences performance, yet the relationship is non-linear: substantially larger models do not guarantee proportionally better results. Input quality, particularly the structural preservation of OCR output, emerges as a critical factor independent of downstream model capability. These findings provide practical guidance for practitioners selecting models for document processing workflows while highlighting fundamental limitations in current open-source approaches to structured information extraction. 
\end{abstract}

\section{Introduction}
The ability to extract structured information from unstructured or semi-structured sources has become an increasingly critical capability in modern data-driven workflows. Digital transformations across all sectors require these workflows to reliably integrate information from scanned documents or digitally edited forms. The rapid advancement of Optical Character Recognition (OCR), Large Language Models (LLMs), and Vision-Language Models (VLMs) has fundamentally transformed this landscape, making such extraction tasks more accurate, scalable, and accessible than ever before.
Next to proprietary solutions, often marketed as all-encompassing general-purpose models, a rich ecosystem of free-to-use, open-source OCR engines, LLMs, and VLMs has emerged. These openly available models present substantial opportunities for processing large volumes of unstructured data, such as the ubiquitous PDF format, without incurring the costs or restrictions of commercial systems.

Leveraging this potential requires more and systematic empirical evaluations of how freely available, off-the-shelf models perform on realistic extraction tasks. This is even more important given that most of this technology has been developing faster than corresponding regulation. While many benchmarks and evaluations focus on tasks that are either simple or not classified as \textit{high-risk} according to the EU AI Act \cite{EUAIAct2024}, using the technology in high risk settings such as health care or workflows in the public sector requires extensive empirical evaluations. Even though these kinds of evaluations are more on the applied side of research we argue that such evaluations are contributions that can and should be made by researchers to build trust in open source technology in the public sector. 

\begin{figure}[H]
\includegraphics[width=\textwidth]{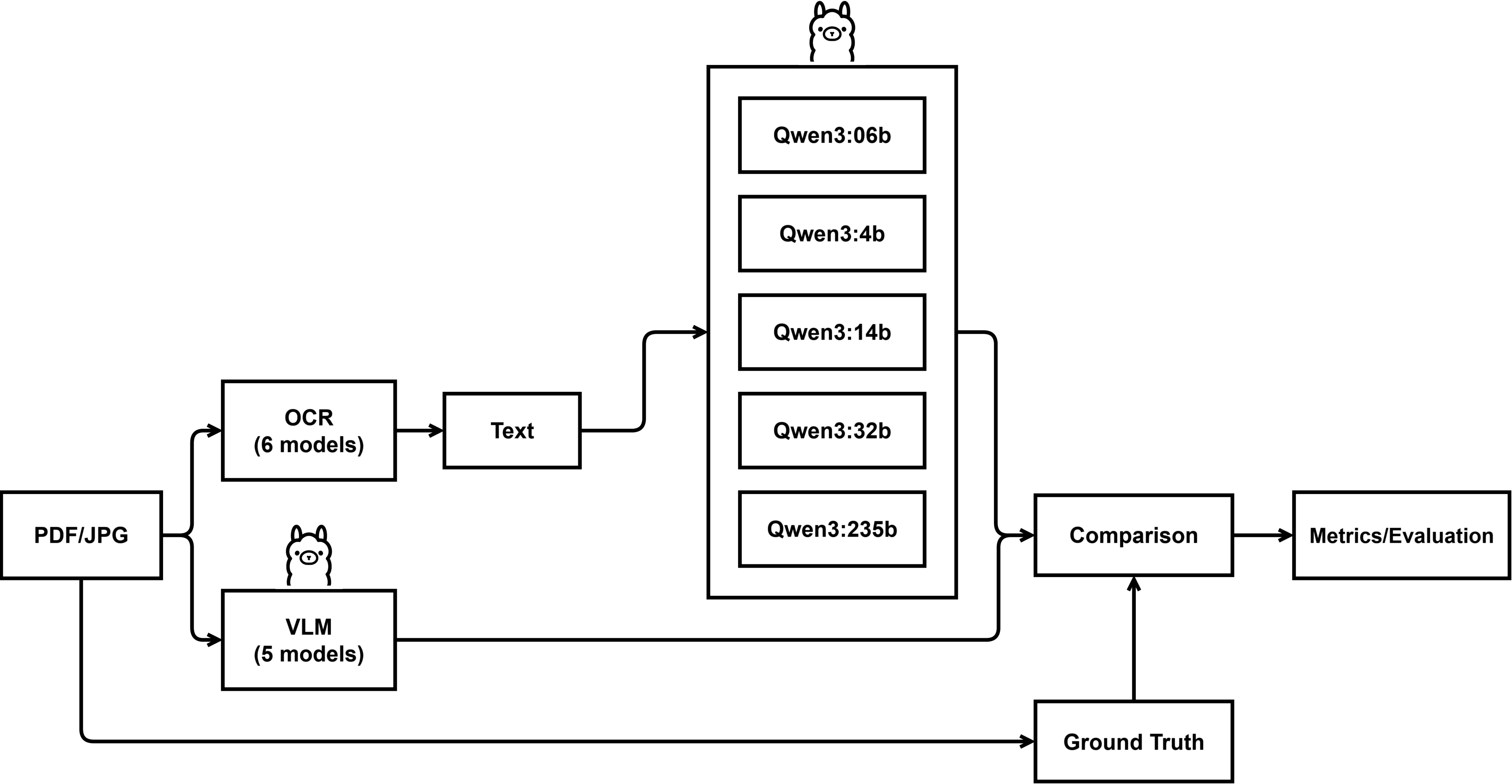}
\caption{Workflow of extraction and evaluation pipeline}
\label{benchmark_procedure}
\end{figure}

To address this challenge, this work makes two main contributions. For one, we devise an evaluation protocol for extraction performance of a high risk use case in the public sector, student applications in an international study program. This task is representative for many other complex tasks: it requires processing of heterogeneous PDF documents containing a mixture of structured and unstructured data, extracting targeted information, and returning it in a well-defined structured format. This workflow, illustrated in Figure~\ref{benchmark_procedure} is representative of operational use cases encountered across numerous domains. Second we use this evaluation protocol for a benchmark assessing the capabilities of open OCR systems combined with LLMs, as well as standalone VLMs. The results are discussed with respect to the extraction performance as well as other factors relevant for practical usage, such as computational efficiency. 

\section{Related Work}
While OCR engines, LLMs, and VLMs have each been evaluated extensively in single-step scenarios, benchmarks assessing their individual or combined performance on complex, multi-step extraction tasks in a zero-shot setting remain sparse. Such tasks, which involve converting documents into machine-readable text, extracting specified information, and returning it in a strictly defined structure, closely reflect how these models are likely to be deployed in practice and constitute precisely the end-to-end pipeline evaluated in this work.
Existing benchmarks tend to isolate individual components of this pipeline. 

With respect to OCR, Khan et al.~\cite{hadi2024benchmarking} evaluated several engines on the CBC Reports Dataset, measuring accuracy, processing time and error rates. PaddleOCR and EasyOCR emerged as the strongest performers, a finding that directly informed the selection of OCR models used in this work.
On the side of information extraction from unstructured and semi-structured documents, Ntinopoulos et al.~\cite{ntinopoulos2025llm} benchmarked mostly LLMs and table transformers on electronic health records in a zero-shot setting, finding that proprietary models consistently achieved strong results with F1 scores above 0.9, outperforming several open-source models. They also found that table transformers achieved higher extraction performance than LLMs when only layout structure is considered, but LLMs surpass table transformers once cell content becomes relevant. Similarly, Gao et al.~\cite{gao2023benchmarking} examined fine-grained extraction across multiple information types, observing that encoder-decoder architectures generalize better to unseen information types, while decoder-only architectures handle unseen task forms more effectively. Notably, model scale did not consistently predict performance, a theme that recurs across the literature.

The interplay between OCR and multimodal LLMs for table extraction was investigated by Nunes et al.~\cite{nunes2025benchmarking}, who benchmarked OCR systems and MLLMs on extracting tables from images using a subset of non-complex tables from PubTables-1M. Although their setup does not include a full end-to-end extraction pipeline, it closely aligns with the benchmark presented here and provides valuable insights. Similar to this work, they enforce structured outputs by prompting models with a predefined Pydantic schema. Evaluating both proprietary and open models, they report strong performance across categories, with several models achieving F1 scores between 85 and 95, depending on the evaluation aspect.

Regarding structured output generation, Tenckhoff et al.~\cite{tenckhoff2026llmstructbench} benchmarked 22 open-source models across five prompting strategies on a synthetically generated dataset, evaluating their ability to extract structured data and produce valid JSON from natural-language text, with GPT-4o included as a proprietary reference. The results reinforce that prompting strategy and model architecture are at least as influential as model scale, and that leading open-source models such as Gemma3-27B can rival proprietary systems.

Finally, Roberts et al.~\cite{roberts2024image2struct} assessed VLMs on their ability to extract structure directly from images, using an automatic evaluation approach based on rendering model outputs and comparing them to the original. Closed-weight models, in particular GPT-4o, significantly outperformed open-weight alternatives, and VLMs struggle to pick up visual nuances and were shown to be sensitive to prompt formulation, suggesting that zero-shot standardized prompting may be suboptimal for such tasks.

Taken together, these works highlight meaningful progress across the individual components relevant to document extraction pipelines, while also underscoring the absence of benchmarks that evaluate free and open models holistically on the complete three-step problem of document ingestion, information extraction and structured output generation that this paper directly addresses.

\section{Methods}
In the following we first describe the study program application data and the extraction task, followed by a description of the extraction workflow and evaluation metrics. 
\subsection{Data}
\begin{figure}[H]
\centering
\includegraphics[width=0.8\textwidth]{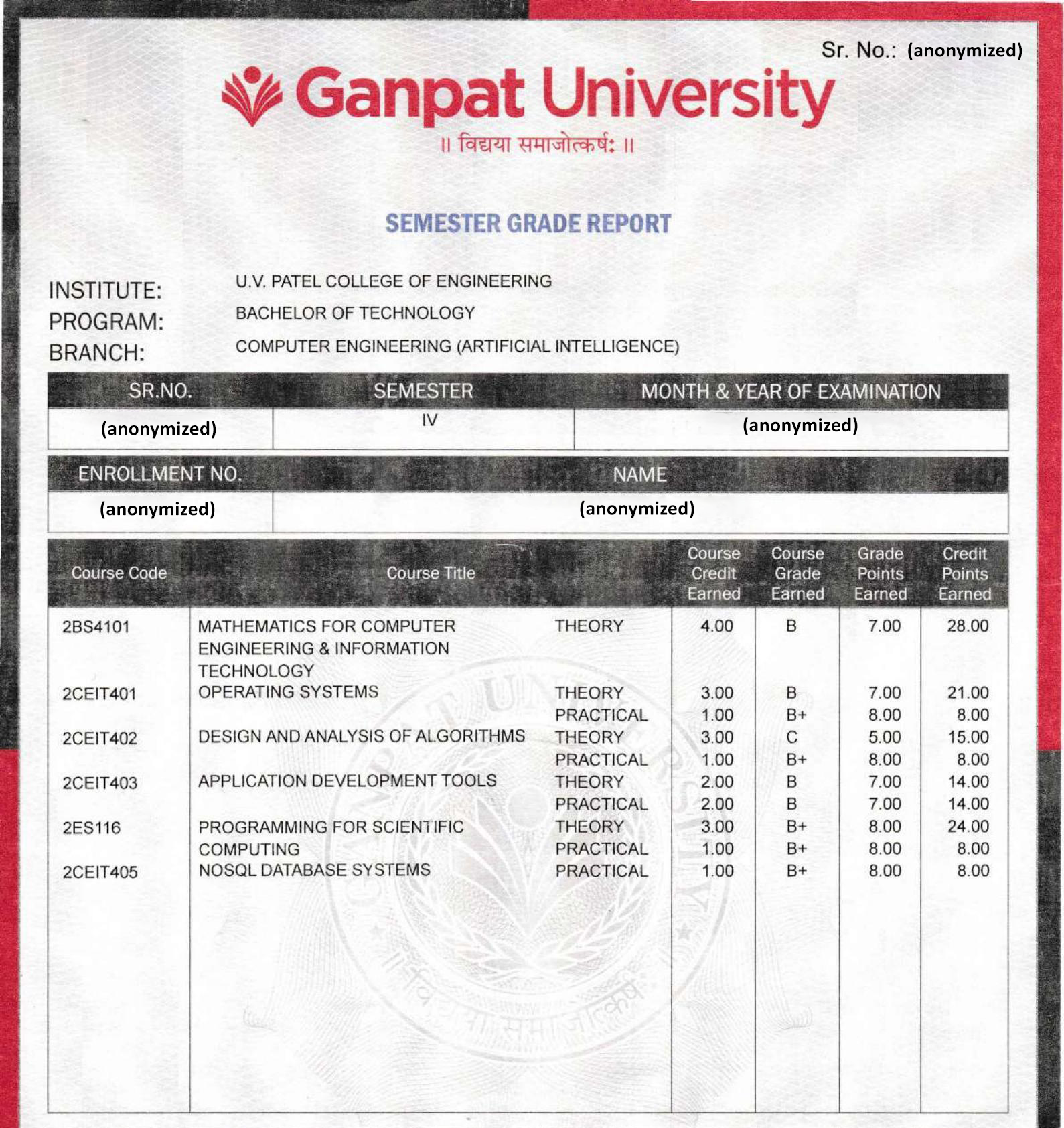}
\caption{Anonymized example document.}
\label{exampledoc}
\end{figure}

The dataset used in this benchmark consists of 100 academic transcripts submitted as part of applications to a German Data Science Master's programme, provided as PDF documents. Of these, 94 are written entirely in English, three are bilingual in English and Turkish, two are in German, and one combines English and Russian. The vast majority of applicants originate from Asian countries, accounting for over 90\% of the dataset. Each document typically contains structured data in the form of tables as well as unstructured elements such as free text, images and institutional sigils. As the documents contain personal information, the dataset cannot be published in its current form and must be anonymized prior to any release. An example document is shown in Figure~\ref{exampledoc}.

All documents were standardized to A4 format and additionally converted to JPG images to accommodate models requiring image input. Each document was then manually parsed to extract all courses deemed relevant to the extraction task, yielding a ground truth against which model outputs are evaluated. The extraction follows a ruleset designed to handle the heterogeneity and formatting inconsistencies typical of international academic documents:

\begin{itemize}
\item Extract only Computer Science or Mathematics courses.
\item In multi-language documents, prefer the English version.
\item In case of multiple grade formats, prefer letter grades.
\item Missing values are filled with \textit{N/A}.
\item Without a clear designation of total credits, select the first value in a row.
\item Formatting inconsistencies such as erroneous spacing or incorrect characters are preserved as-is in the ground truth.
\end{itemize}

Table~\ref{examplegt}, which shows the ground truth belonging to the document shown in Figure~\ref{exampledoc}, illustrates the application of these rules. With multiple credit columns given and none being explicitly designated, the first credit value for each course is selected. For example, "OPERATING SYSTEMS" appears with both theory (3.0) and practical (1.0) credits in the original document, but only 3.0 is extracted as it is the first credit value associated with that course.

The extracted ground truth comprises four columns: \textit{academic field}, \textit{course name}, \textit{grade}, and \textit{awarded credits}, totaling 952 unique rows across all documents. The number of extracted rows per document ranges from 0 to 44, with no duplicate entries within any single document.

\begin{table}[H]
\caption{Manually extracted Ground Truth for the Document in Figure~\ref{exampledoc}}
\label{examplegt}
\begin{tabular}{|l|p{6cm}|l|l|}
\hline
\textbf{Academic Field} & \textbf{Course Name} & \textbf{Grade} & \textbf{Awarded Credits} \\
\hline
Mathematics     & MATHEMATICS FOR COMPUTER THEORY ENGINEERING \& INFORMATION TECHNOLOGY & B  & 4.00 \\
Computer Science & OPERATING SYSTEMS                                                      & B  & 3.00 \\
Computer Science & DESIGN AND ANALYSIS OF ALGORITHMS                                      & C  & 3.00 \\
Computer Science & APPLICATION DEVELOPMENT TOOLS                                          & B  & 2.00 \\
Computer Science & PROGRAMMING FOR SCIENTIFIC THEORY COMPUTING                            & B+ & 3.00 \\
Computer Science & NO SQL DATABASE SYSTEMS                                                & B+ & 1.00 \\
\hline
\end{tabular}
\end{table}

\subsection{Extraction}
We compare the performance of OCR engines combined with LLMs against standalone VLMs in a three-step pipeline: digitalizing and preparing document content, extracting task-relevant information as specified in the prompt, and returning it as a structured output. All LLMs and VLMs are run using the Ollama platform, which provides consistent base settings across all models as well as an integrated format parameter that enforces structured output via a Pydantic schema.
The OCR engines used in this benchmark were selected based on two criteria: popularity among practitioners, measured by GitHub star count, and performance reported in recent benchmarking literature~\cite{heakl2025kitabb}~\cite{nazeem2024open}. All models are used with pretrained weights without any further fine-tuning. The following engines were evaluated: Pytesseract~\cite{smith2007overview}, PP-OCRv5 (PaddleOCR)~\cite{cui2025paddleocr}, PP-StructureV3 (PaddleOCR)~\cite{cui2025paddleocr}, MinerU~\cite{wang2024mineru}, EasyOCR~\cite{easyocr2020}, and docTR~\cite{liao2023doctr}. Table~\ref{tab:ocr} provides an overview of each model's input, output and language configuration. Table~\ref{tab:ocr-arch} details the architecture components for engines where explicit configuration was specified, with selections based on the highest-performing options as indicated in each engine's official documentation and benchmarks.

\begin{table}[H]
\caption{OCR model configurations and output formats.}
\label{tab:ocr}
\begin{tabularx}{\textwidth}{|l|l|l|X|}
\hline
\textbf{Model} & \textbf{Input} & \textbf{Language Config} & \textbf{Output Format}  \\
\hline
docTR         & PDF         & auto-detect & Unstructured; line breaks between text blocks \\
EasyOCR       & Image (JPG) & English     & Unstructured; tabs between text blocks (manually added) \\
MinerU        & PDF         & English     & Structured; HTML tags for tables, line breaks between text rows \\
PPOCRv5       & Image (JPG) & English     & Unstructured; tabs between text blocks (manually added) \\
PPStructureV3 & PDF         & English     & Structured; HTML tags for tables, line breaks between text blocks \\
Pytesseract   & Image (JPG) & English     & Structured; line breaks between text rows \\
\hline
\end{tabularx}
\end{table}

\begin{table}[H]
\caption{OCR model architecture components (specified models only).}
\label{tab:ocr-arch}
\resizebox{\textwidth}{!}{%
\begin{tabular}{|l|l|l|l|l|l|}
\hline
\textbf{Model} & \textbf{Orient. Class.} & \textbf{Doc Unwarping} & \textbf{Text Detection} & \textbf{Textline Orient.} & \textbf{Text Recognition}\\
\hline
docTR         & ---                                & ---     & \texttt{db\_resnet50}       & ---                                     & \texttt{master} \\
EasyOCR       & ---                                & ---     & ---                         & ---                                     & --- \\
MinerU        & ---                                & ---     & ---                         & ---                                     & --- \\
PPOCRv5       & \texttt{PP-LCNet\_x1\_0\_doc\_ori} & \texttt{UVDoc} & \texttt{OCRv5\_server\_det} & \texttt{PP-LCNet\_x1\_0\_textline\_ori} & \texttt{PP-OCRv5\_server\_rec} \\
PPStructureV3 & \texttt{PP-LCNet\_x1\_0\_doc\_ori} & \texttt{UVDoc} & \texttt{OCRv5\_server\_det} & \texttt{PP-LCNet\_x1\_0\_textline\_ori} & \texttt{PP-OCRv5\_server\_rec} \\
Pytesseract   & ---                                & ---     & ---                         & ---                                     & --- \\
\hline
\end{tabular}%
}
\end{table}

For the extraction of relevant structured information from the OCR output, a single LLM was deliberately chosen to reduce variables and place the focus on the effect of model scale. Qwen3~\cite{yang2025qwen3}, a modern model family released in 2025, was selected as it offers a wide range of parameter counts within a single architecture. Five parameter sizes were evaluated, as detailed in Table~\ref{tab:qwen3}.

\begin{table}[H]
\centering
\caption{Qwen3 model architecture parameters~\cite{yang2025qwen3}. Context lengths as specified in Ollama~\cite{ollama2025qwen3}. Q/KV heads refer to query and key-value attention heads~\cite{vaswani2023attention}.}
\label{tab:qwen3}
\begin{tabular}{|l|l|l|l|l|}
\hline
\textbf{Model} & \textbf{Parameters} & \textbf{Context Length} & \textbf{Layers} & \textbf{Heads (Q / KV)}\\
\hline
Qwen3 & 0.6B & 40K  & 28 & 16 / 8\\
Qwen3 & 4B   & 256K & 36 & 32 / 8\\
Qwen3 & 14B  & 40K  & 40 & 40 / 8\\
Qwen3 & 32B  & 40K  & 64 & 64 / 8\\
Qwen3 & 235B & 256K & 94 & 64 / 4\\
\hline
\end{tabular}
\end{table}

\subsubsection{VLM based Extraction}

VLMs were selected following the similar criteria as the OCR engines, drawing from both the most used models from Ollama model zoo and recent literature~\cite{li2025survey}. An effort was made to keep parameter counts comparable across models at approximately 30 billion parameters, though not all models offered variants at this scale. The following models were evaluated: LLaVA (7B)~\cite{liu2023visual}, Ministral-3 (14B)~\cite{liu2026ministral} , Gemma3 (27B)~\cite{gemmateam2025gemma}, Qwen2.5VL (32B)~\cite{bai2025qwen25vl}, and Qwen3VL (32B)~\cite{bai2025qwen3vl}. As with the LLM based pipeline, structured outputs were generated using the same Pydantic schema via the Ollama format API. Table~\ref{tab:vlm} provides a detailed overview of each model.

\begin{table}[H]
\centering
\caption{Vision-language model overview.}
\label{tab:vlm}
\begin{tabular}{|l|l|l|l|l|l|}
\hline
\textbf{Model} & \textbf{Year} & \textbf{Architecture} & \textbf{Parameters} & \textbf{Context} & \textbf{Vision Encoder}\\
\hline
Gemma 3     & 2025 & Decoder-only    & 27B & 128K & SigLIP \\
LLaVA       & 2024 & Encoder-decoder & 7B  & 32K  & CLIP \\
Ministral 3 & 2025 & Decoder-only    & 14B & 256K & ViT \\
Qwen2.5-VL  & 2025 & Encoder-decoder & 32B & 125K & ViT \\
Qwen3-VL    & 2025 & Encoder-decoder & 32B & 256K & SigLIP-2 \\
\hline
\end{tabular}
\end{table}

\subsubsection{Prompting Strategy and Experimental Design}

The prompt design incorporates several evidence-based strategies from recent literature: meaning-typed prompting via Pydantic schemas rather than rigid JSON definitions for reliable structured output~\cite{irugalbandara2024meaning}, a verification phase where the model reviews its own output~\cite{polak2024extracting}, delimiters to separate schema definitions and text sections~\cite{chen2025evaluation}, and concise, direct instructions~\cite{li2023practical}.
Structured output is enforced via the following Pydantic schema:

\begin{lstlisting}[language=Python, basicstyle=\small\ttfamily, frame=single]
class Course(BaseModel):
    academic_field: Literal["Computer Science", "Mathematics"]
    course_name: str
    grade: float | str | None = None
    awarded_credits: float | None = None

class Courses(BaseModel):
    courses: List[Course]
\end{lstlisting}

The prompt follows a structured format: input data and goal definition, schema presentation with illustrative examples, output rules, context provision, and a verification section with output format specifications. Temperature is set to 0 for all models, following best practices for deterministic structured generation as recommended in the Ollama documentation~\cite{ollama2025structured}.

All experiments were executed on a compute cluster using a single NVIDIA H200 GPU with 141 GB VRAM, 32 GB system memory, and 8 CPU cores. System memory and CPU utilization remained well below capacity throughout all experiments. For the Qwen3:235b model, an NVIDIA B200 GPU with 192 GB VRAM was briefly utilized to accelerate inference.

\subsection{Evaluation}

%
Model outputs were first validated and normalized to ensure comparability with ground truth. Empty grade and credit cells columns were filled with \texttt{N/A} to match the ground truth format. When models failed to return data in the required structure due to recursive reasoning patterns, additional text generation, inclusion of extra columns, or omission of required fields, an empty dataframe with the correct schema was substituted, effectively treating non-compliant outputs as complete extraction failures.

Both the extracted data and ground truth underwent identical normalization procedures: (1) index columns were dropped, (2) leading and trailing whitespace was stripped, (3) academic field and course name values were converted to lowercase, and (4) numerical values were cast to floats, while any alphabetic characters in grade or credit columns were also converted to lowercase.

\subsubsection{Evaluation Metrics}

\begin{figure}[H]
\includegraphics[width=\textwidth]{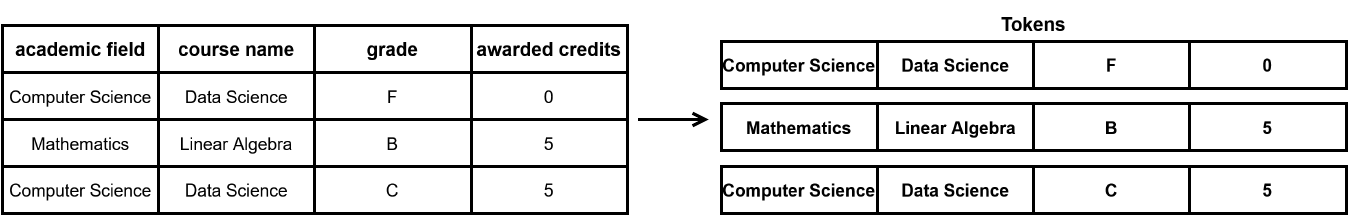}
\caption{Tokenization visualized}
\label{tokenization}
\end{figure}

For our evaluation we focus on exact string matches after lowercasing.
%
Both the ground truth and extracted data were tokenized by converting each row, consisting of academic field, course name, grade, and awarded credits, into a single unique token as is illustrated in Figure~\ref{tokenization}. Given that ground truth contains no duplicate rows, both datasets were represented as sets of row tokens, enabling efficient set-based comparison. From these sets, a combined set of all unique tokens was constructed using set intersection.

Following standard practice in information or text extraction benchmarks~\cite{adhikari2025comparative,kim2011figure,nasar2018information} we computed Precision, Recall, and F1 scores.
%
%
To determine True Positives (TP), False Positives (FP), and False Negatives (FN) for these metrics, we adopt a set-theoretic approach based on Jaccard similarity~\cite{adhikari2025comparative}:
%
\begin{itemize}
    \item \textbf{True Positives (TP)}: Tokens present in both extracted output and ground truth
    \item \textbf{False Positives (FP)}: Tokens in extracted output but absent from ground truth
    \item \textbf{False Negatives (FN)}: Tokens in ground truth but absent from extracted output
\end{itemize}
In the subsequent analysis, F1 Score serves as the primary metric for model evaluation and comparison, as it provides a balanced measure of both extraction accuracy and completeness, making it particularly suitable for assessing performance on imbalanced extraction tasks where both false positives and false negatives carry significant cost.

\subsection{Results}
\begin{figure}
\includegraphics[width=\textwidth]{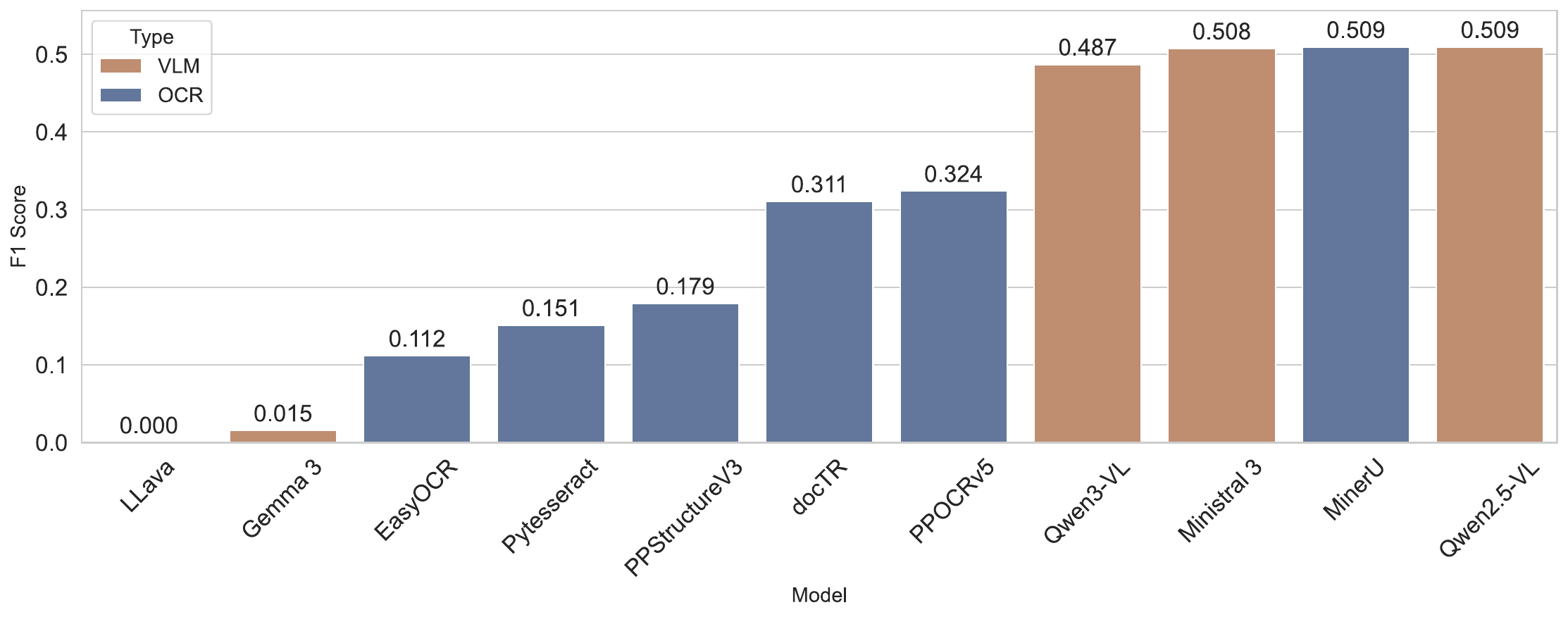}
\caption{Best F1 score per model: top OCR+LLM combination (always Qwen3:235B) versus all VLMs.}
\label{bestf1}
\end{figure}

In Table~\ref{tab:results} we show the overall results for each OCR+LLM and VLM model and Figure \ref{bestf1} shows the best F1 score per model, comparing the top-performing OCR+LLM combination for each OCR engine against all VLMs. 

\paragraph{VLMs outperform OCR+LLM approaches} In the OCR+LLM pipeline, the best-performing LLM was Qwen3:235B in all cases. The results reveal a clear performance gap between OCR engines, largely attributable to the quality and structure of their output. MinerU stands out as the strongest OCR engine, achieving an F1 score of 0.509, placing it on par with the best-performing VLM, Qwen2.5-VL (0.509). This can be attributed to MinerU's output format, which preserves the spatial layout of the page by grouping text that appears on the same horizontal line into a single row, and representing tabular content using HTML tags to explicitly encode cell boundaries. Other OCR engines, by contrast, insert line breaks after each individual text block regardless of its position on the page, and largely fail to represent table structure semantically. This directly demonstrates the critical importance of input quality to downstream LLM performance: the structure of the OCR output is at least as consequential as the capability of the LLM processing it.
On the VLM side, results are polarized. Qwen2.5-VL (0.509) and Ministral-3 (0.508) achieve the highest scores overall, while LLaVA (0.000) and Gemma3 (0.015) perform at or near zero, most likely due to malformed outputs that were treated as complete extraction failures under the evaluation protocol. Notably, Qwen2.5-VL marginally outperforms its successor Qwen3-VL (0.487), suggesting that a newer model version does not necessarily translate to improved task-specific performance. Overall, results across both pipelines are poor: the mean F1 score across all VLMs and the best OCR+LLM combination per engine is 0.282, with the highest observed score capped at 0.509.
\begin{table}
\centering
\caption{Extraction performance metrics for OCR+LLM pipeline and VLM models.}
\label{tab:results}
\begin{tabular}{|l|l|r|r|r|}
\hline
\textbf{OCR Engine} & \textbf{LLM Model} & \textbf{Precision} & \textbf{Recall} & \textbf{F1} \\
\hline
\multicolumn{5}{|c|}{\textit{OCR + LLM Pipeline}} \\
\hline
docTR & Qwen3 0.6B & 0.014 & 0.022 & 0.016 \\
docTR & Qwen3 4B & 0.199 & 0.217 & 0.201 \\
docTR & Qwen3 14B & 0.202 & 0.238 & 0.215 \\
docTR & Qwen3 32B & 0.204 & 0.222 & 0.209 \\
docTR & Qwen3 235B & \textbf{0.301} & \textbf{0.329} & \textbf{0.311} \\
\hline
EasyOCR & Qwen3 0.6B & 0.007 & 0.009 & 0.008 \\
EasyOCR & Qwen3 4B & 0.079 & 0.088 & 0.080 \\
EasyOCR & Qwen3 14B & 0.079 & 0.111 & 0.088 \\
EasyOCR & Qwen3 32B & 0.103 & 0.125 & 0.108 \\
EasyOCR & Qwen3 235B & \textbf{0.104} & \textbf{0.126} & \textbf{0.112} \\
\hline
MinerU & Qwen3 0.6B & 0.078 & 0.081 & 0.075 \\
MinerU & Qwen3 4B & 0.415 & 0.472 & 0.427 \\
MinerU & Qwen3 14B & 0.455 & 0.499 & 0.465 \\
MinerU & Qwen3 32B & 0.471 & \textbf{0.551} & 0.501 \\
MinerU & Qwen3 235B & \textbf{0.489} & 0.542 & \textbf{0.509} \\
\hline
PPOCRv5 & Qwen3 0.6B & 0.002 & 0.002 & 0.002 \\
PPOCRv5 & Qwen3 4B & 0.160 & 0.199 & 0.170 \\
PPOCRv5 & Qwen3 14B & 0.207 & 0.262 & 0.222 \\
PPOCRv5 & Qwen3 32B & 0.163 & 0.184 & 0.169 \\
PPOCRv5 & Qwen3 235B & \textbf{0.312} & \textbf{0.345} & \textbf{0.324} \\
\hline
PPStructureV3 & Qwen3 0.6B & 0.022 & 0.020 & 0.020 \\
PPStructureV3 & Qwen3 4B & 0.132 & 0.146 & 0.129 \\
PPStructureV3 & Qwen3 14B & 0.123 & 0.133 & 0.124 \\
PPStructureV3 & Qwen3 32B & 0.157 & 0.163 & 0.157 \\
PPStructureV3 & Qwen3 235B & \textbf{0.176} & \textbf{0.191} & \textbf{0.179} \\
\hline
Pytesseract & Qwen3 0.6B & 0.005 & 0.005 & 0.005 \\
Pytesseract & Qwen3 4B & 0.091 & 0.095 & 0.092 \\
Pytesseract & Qwen3 14B & 0.086 & 0.100 & 0.091 \\
Pytesseract & Qwen3 32B & 0.090 & 0.096 & 0.092 \\
Pytesseract & Qwen3 235B & \textbf{0.145} & \textbf{0.162} & \textbf{0.151} \\
\hline
\multicolumn{5}{|c|}{\textit{VLM-Only Pipeline}} \\
\hline
\multicolumn{2}{|l|}{\textbf{VLM Model}} & \textbf{Precision} & \textbf{Recall} & \textbf{F1} \\
\hline
\multicolumn{2}{|l|}{Gemma3 27B} & 0.021 & 0.018 & 0.015 \\
\multicolumn{2}{|l|}{LLaVA 7B} & 0.000 & 0.000 & 0.000 \\
\multicolumn{2}{|l|}{Ministral-3 14B} & \textbf{0.480} & 0.565 & 0.508 \\
\multicolumn{2}{|l|}{Qwen2.5-VL 32B} & 0.473 & \textbf{0.574} & \textbf{0.509} \\
\multicolumn{2}{|l|}{Qwen3-VL 32B} & 0.441 & 0.573 & 0.487 \\
\hline
\end{tabular}
\end{table}

\begin{figure}[H]
\includegraphics[width=\textwidth]{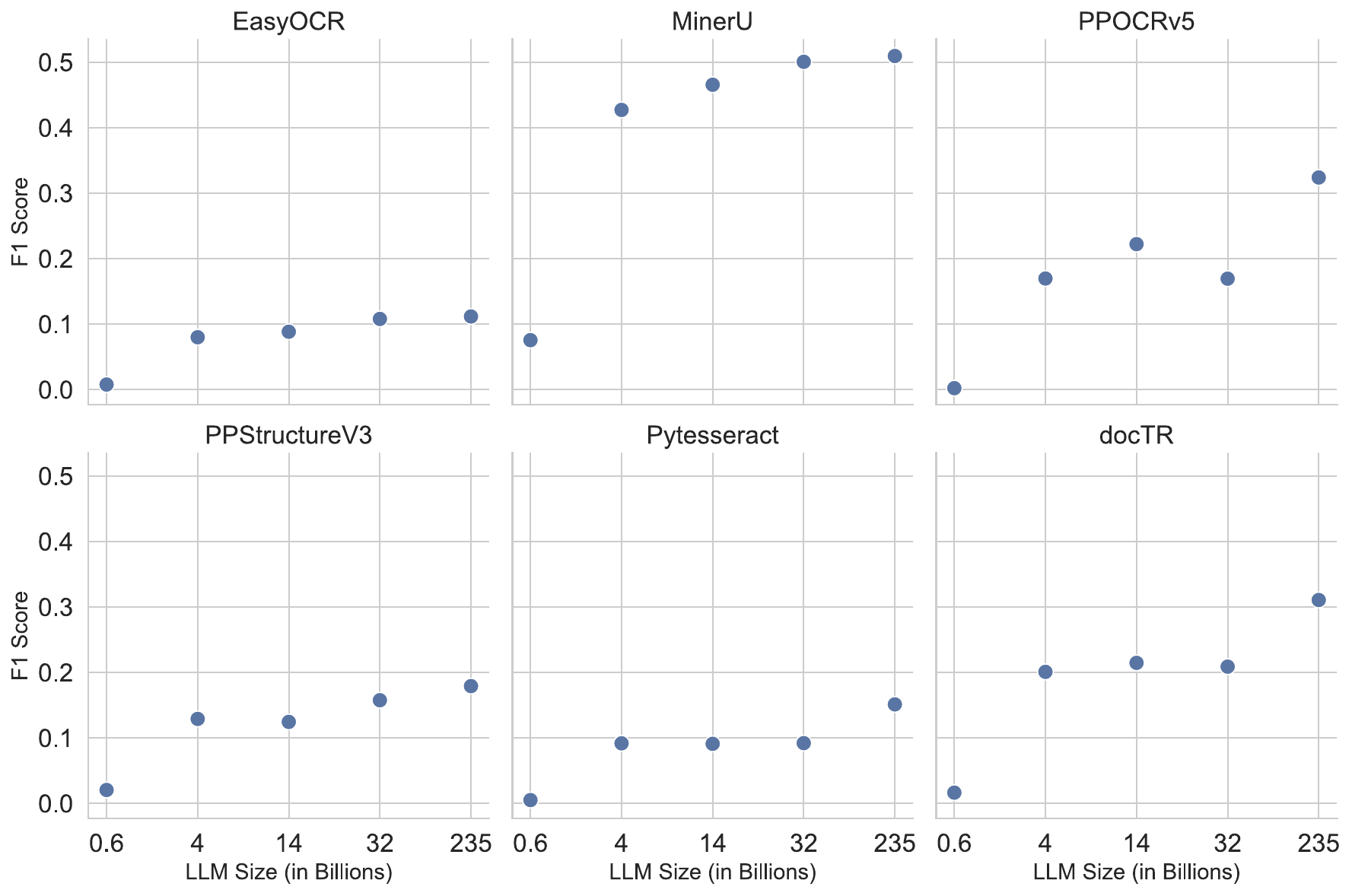}
\caption{F1 scores across Qwen3 parameter sizes for each OCR engine.}
\label{parameters}
\end{figure}

\paragraph{Parameter Size and Extraction Performance}
Figure~\ref{parameters} illustrates how F1 score varies across Qwen3 parameter sizes for each OCR engine. A consistent pattern emerges: the 0.6B model performs substantially worse than all larger variants across every OCR engine, frequently producing near-zero scores. This is likely not solely a function of model capacity, but also a known instability of the Qwen3:0.6B model, which has been documented to enter a repetitive generation loop, producing incoherent output that fails to conform to the required Pydantic schema. Beyond the 0.6B outlier, larger models generally yield modest but consistent improvements, with the 235B variant typically achieving the highest score per engine. However, this trend is not universal: for PPOCRv5 and docTR, the 14B model scores marginally higher than the 32B model, reinforcing the finding from related work that model scale and task performance do not maintain a strictly linear relationship.

\begin{figure}
\includegraphics[width=\textwidth]{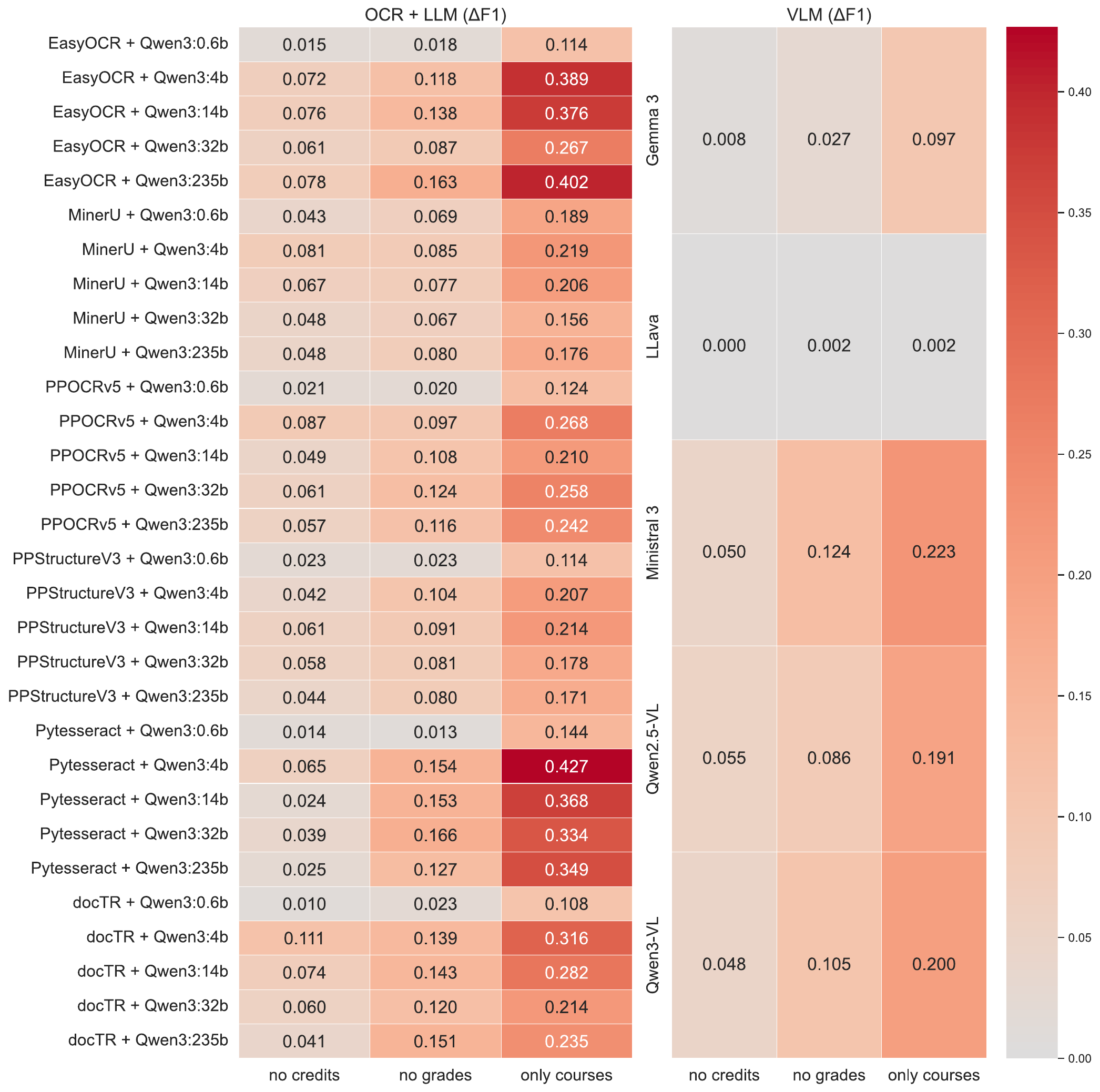}
\caption{F1 improvement when removing grade, awarded credits, or both.}
\label{f1delta}
\end{figure}

\paragraph{Simplified Extraction with Fewer Attributes}
To investigate the sensitivity of the performance with reduced task complexity, we simplified the task and neglected the grade and awarded credits attributes. Figure~\ref{f1delta} presents the improvement in F1 score ($\Delta$F1) when the grade column, the awarded credits column, or both are removed from both ground truth and model output prior to evaluation. Positive deltas indicate improved scores under the reduced column setting. Across all model configurations, the largest improvements are observed when both columns are removed, retaining only academic field and course name. 
%
Removing grades consistently yields a larger improvement than removing awarded credits, though the magnitude varies considerably across configurations, ranging from below 0.01 to above 0.1. This suggests that grade extraction poses a greater challenge, likely due to the diversity of grading formats across transcripts, including both letter and numerical grades, which introduces an additional layer of ambiguity even when explicit handling rules are provided in the prompt. These patterns hold consistently across both VLM and OCR+LLM configurations.
Finally, as observed in Figure~\ref{parameters}, the 0.6B Qwen3 configurations show minimal improvement even when columns are removed. This further supports the interpretation that the near-zero baseline scores for these configurations are driven primarily by output invalidity due to the model's repetitive generation behavior, rather than extraction difficulty.

\section{Conclusion}

This benchmark evaluated the end-to-end capability of open-source OCR engines, LLMs, and VLMs to extract structured information from heterogeneous real-world documents in a zero-shot setting, using all models in their pretrained state with best available configurations. The results reveal that none of the tested approaches can be considered a reliable out-of-the-box solution for complex extraction tasks of this kind, with roughly 75\% of all configurations scoring below an F1 of 0.25.

A central finding is that the quality of LLM output is strongly dependent on the quality of the context provided. OCR engines that preserve the spatial and structural layout of the source document, as MinerU does via row-wise text grouping and HTML-encoded tables, yield dramatically better downstream extraction results than engines that produce unstructured, block-fragmented output. This underlines that input preparation is at least as important as model capability in such pipelines.

VLMs generally outperform OCR+LLM pipelines in this task, as they bypass the OCR step entirely and operate directly on the document image. However, their performance is far from reliable, with several models producing near-zero scores due to malformed outputs. It is also worth noting that a newer model is not always a better one: Qwen2.5-VL outperformed its successor Qwen3-VL on this task, and similarly, larger LLM parameter counts do not guarantee proportionally better results, though very small models such as Qwen3:0.6B are clearly insufficient for tasks of this complexity.

These findings apply specifically to the zero-shot, off-the-shelf setting evaluated here. The performance levels observed are not necessarily an inherent ceiling: improved preprocessing and data quality feeding into the LLM or VLM, more targeted prompting strategies, and refined schema definitions are all promising directions that could meaningfully improve results. These do however require additional effort and experimentation, and suggest that deploying such pipelines reliably in practice demands more than simply selecting a capable model off the shelf. Improved data preparation, more targeted prompting strategies, and refined schema definitions are all promising directions that could meaningfully improve performance without requiring fine-tuning, and warrant further investigation.

\subsubsection{\discintname}
The authors have no competing interests to declare that are relevant to the content of this article.

%
%
\bibliographystyle{splncs04}
\bibliography{ref}

\end{document}